\documentclass[10pt,journal,compsoc]{IEEEtran}
\ifCLASSOPTIONcompsoc
  \usepackage[nocompress]{cite}
\else
  \usepackage{cite}
\fi
\ifCLASSINFOpdf
\else
\fi

\usepackage{graphicx}
\usepackage{amsmath}
\usepackage{amssymb}

\begin{document}
%

\title{Pseudo-LiDAR Point Cloud Interpolation Based on 3D Motion Representation and Spatial Supervision}
\author{Haojie~Liu, Kang~Liao, Chunyu~Lin,
Yao~Zhao,~\IEEEmembership{Senior Member,~IEEE}, Yulan~Guo
\thanks{\textit{Corresponding author: Chunyu Lin}}
\thanks{Haojie Liu, Kang Liao, Chunyu Lin, Yao Zhao are with the Institute of Information Science, Beijing Jiaotong University, Beijing 100044, China, and also with the Beijing Key Laboratory of Advanced Information Science and Network Technology, Beijing 100044, China (email: hj\_liu@bjtu.edu.cn, kang\_liao@bjtu.edu.cn, cylin@bjtu.edu.cn, yzhao@bjtu.edu.cn).}
\thanks{Yulan Guo, College of Electronic Science and Technology, National University of Defense Technology, yulan.guo@nudt.edu.cn}
}

\IEEEtitleabstractindextext{%
\begin{abstract}
Pseudo-LiDAR point cloud interpolation is a novel and challenging task in the field of autonomous driving, which aims to address the frequency mismatching problem between camera and LiDAR. Previous works represent the 3D spatial motion relationship induced by a coarse 2D optical flow, and the quality of interpolated point clouds only depends on the supervision of depth maps. As a result, the generated point clouds suffer from inferior global distributions and local appearances. To solve the above problems, we propose a Pseudo-LiDAR point cloud interpolation network to generates temporally and spatially high-quality point cloud sequences. By exploiting the scene flow between point clouds, the proposed network is able to learn a more accurate representation of the 3D spatial motion relationship. For the more comprehensive perception of the distribution of point cloud, we design a novel reconstruction loss function that implements the chamfer distance to supervise the generation of Pseudo-LiDAR point clouds in 3D space. In addition, we introduce a multi-modal deep aggregation module to facilitate the efficient fusion of texture and depth features. As the benefits of the improved motion representation, training loss function, and model structure, our approach gains significant improvements on the Pseudo-LiDAR point cloud interpolation task. The experimental results evaluated on KITTI dataset demonstrate the state-of-the-art performance of the proposed network, quantitatively and qualitatively. 
\end{abstract}

}
\maketitle

\IEEEdisplaynontitleabstractindextext

%
\IEEEpeerreviewmaketitle

%
%
%
%
\IEEEraisesectionheading{\section{Introduction}\label{sec:introduction}}
\IEEEPARstart{R}{ecently}, multi-sensor systems that sense both image and depth information have gained increasing attention, which is widely used in navigation applications such as autonomous driving and robotics. For these applications, the accurate and dense depth information is crucial for the obstacle avoidance \cite{yang2019reactive}, object detection \cite{Pointrcnn,fpointnet}, and 3D scene reconstruction tasks \cite{shin20193d}. On the perception platform of autonomous driving, the prerequisite of sensor fusion is the time synchronization of the system. However, there is an inherent limitation in LiDAR sensors, which provide dependable 3D spatial information at a low frequency (around 10Hz). To achieve time synchronization, the frequency of the camera (around 20Hz) has to be decreased, leading to an inefficient multi-sensor system. In addition, LiDAR sensors only obtain sparse depth measurements, e.g. 64 scan lines in the vertical direction. Such a low frequency and sparse depth sensing are insufficient for the actual applications. Therefore, for the synchronous sensing of multi-sensor systems, it would be promising to increase the frequency of LiDAR data to match the high frequency of cameras. The high-frequency and dense point cloud sequences are of great significance in the high-speed and complicated application scenarios. 
\begin{figure}
	\centering
	\includegraphics[width=1\linewidth]{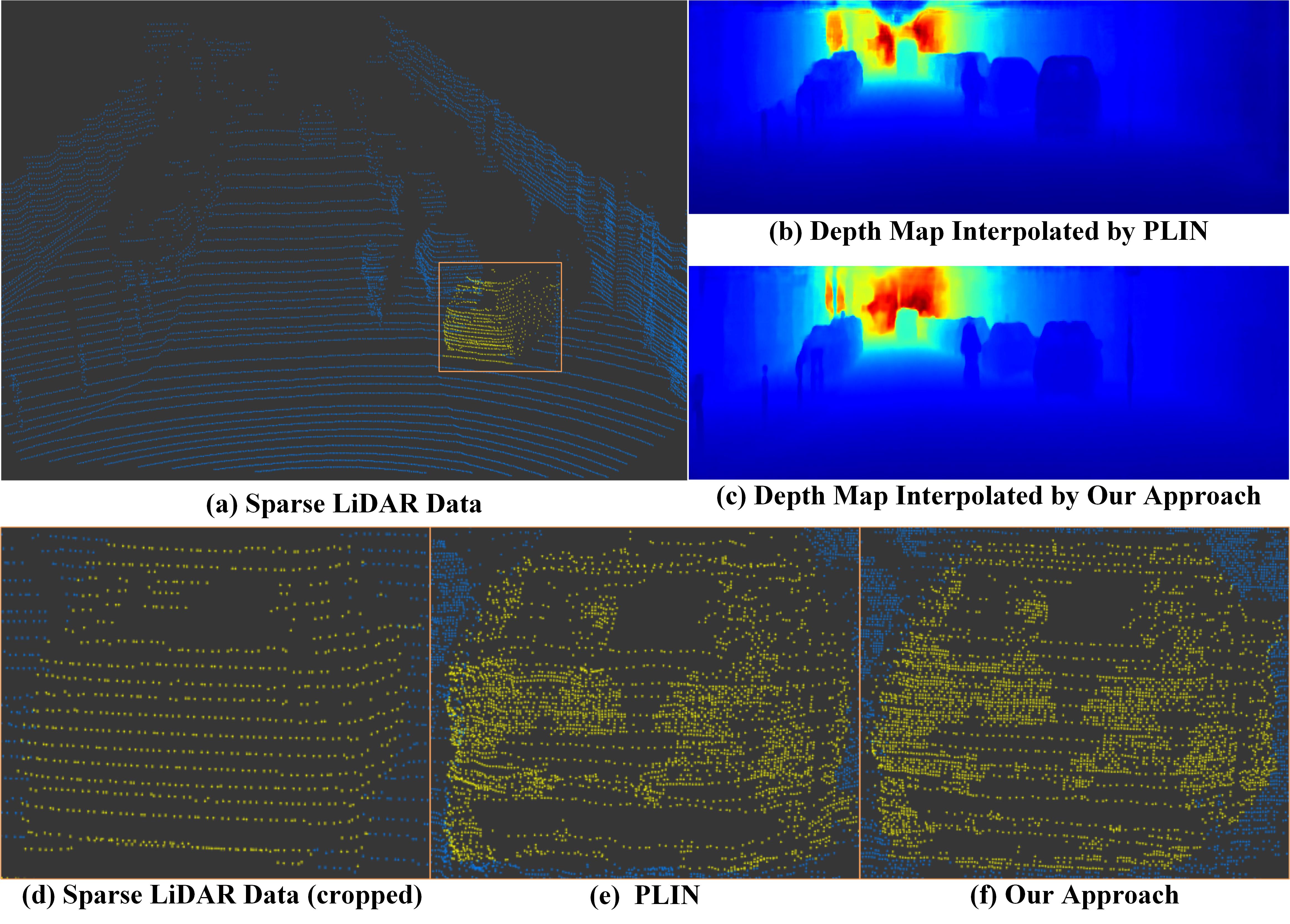}
	\caption{The comparison of Pseudo-LiDAR point cloud interpolation methods. We visualize the interpolated depth maps and Pseudo-LiDAR point clouds obtained by PLIN \cite{liu2020plin} and our approach. The cropped region (d) indicates the ground truth point cloud from the sparse LiDAR data (a). Our result displays more accurate appearances than PLIN and keeps denser distribution than the original point cloud.}
	\label{f1}
\end{figure}

Due to the huge volume of the point cloud captured by LiDAR, directly processing and learning on 3D space is time-consuming. PLIN \cite{liu2020plin} presents the first pipeline for the Pseudo-LiDAR point cloud interpolation task, in which the Pseudo-LiDAR point cloud is obtained by the interpolated dense depth map and camera parameters. PLIN increases the frequency of LiDAR sensors by interpolating adjacent point clouds, to solve the problem of frequency mismatching between LiDAR and camera. Using a coarse-to-fine architecture, PLIN can progressively perceive multi-modal information and generate the intermediate Pesudo-LiDAR point clouds. However, PLIN has several limitations as follows.

1) The spatial motion information is derived from the 2D optical flow between color images of adjacent consecutive frames. Nevertheless, the 2D optical flow only represents the movement deviation of the planar pixels, and cannot represent the movement in 3D space. Thus, the motion relationship used in PLIN causes an inferior temporal interpolation of point cloud sequence.  

2) PLIN only supervises the generation of intermediate depth maps during the training process. Consequently, the quality of generated point clouds only depends on the synthetic depth maps. Moreover, the network does not provide any spatial constraints on the point cloud generation and does not measure the quality of the point cloud.

3) The fusion way of multi-modal features is plain. PLIN roughly concatenates the texture and depth features and feeds these features into an interpolation neural network. However, this type of fusion cannot emphasize the effective complementary message passing between different features.

Based on the above limitations, our work focuses on these challenges. In this paper, we present a novel network to improve the motion representation and spatial supervision for Pseudo-LiDAR point cloud interpolation. In particular, since the optical flow does not describe the motion information in 3D space, we use the scene flow to guide the generation of the Pseudo-LiDAR point cloud. The scene flow represents a 3D motion field from two consecutive stereo pairs, and we design a spatial motion perception module to estimate it. In addition, we implement a point cloud reconstruction loss to constrain the interpolation of the Pseudo-LiDAR point cloud, which enables us to generate more realistic results with respect to the spatial distribution. 

For the architecture of our approach, we design a dual branch structure consisting of texture completion and temporal interpolation. In the texture completion branch, the intermediate color image is used to provide rich textures for the dense depth map generation. In the temporal interpolation branch, we exploit a warping layer with two adjacent point clouds and the estimated scene flow to synthesize the intermediate depth map. To facilitate the efficient fusion of texture and depth features, we introduce a multi-modal deep aggregation module. As the benefits of the improved motion representation, loss function, and model structure, our approach gains significant improvements on the Pseudo-LiDAR point cloud interpolation task. As illustrated in Fig. \ref{f1}, we compare the depth map and point cloud interpolated by PLIN and our approach, and our results display more accurate appearances than PLIN and keep denser distribution than the original.

The contributions of this work are summarized as follows. 
\begin{itemize}
	\item Considering the full representation of 3D motion information, we design a spatial motion perception module to guide the generation of Pseudo-LiDAR point cloud. 
	\item We design a reconstruction loss function to supervise and guide the generation of Pseudo-LiDAR point cloud in 3D space, and further introduce a quality metric of the point cloud. 
	\item We propose a multi-modal deep aggregation module to effectively fuse the feature of the texture completion branch and temporal interpolation branch.
\end{itemize}

\begin{figure*}[t]
	\centering
	\includegraphics[width=1\linewidth]{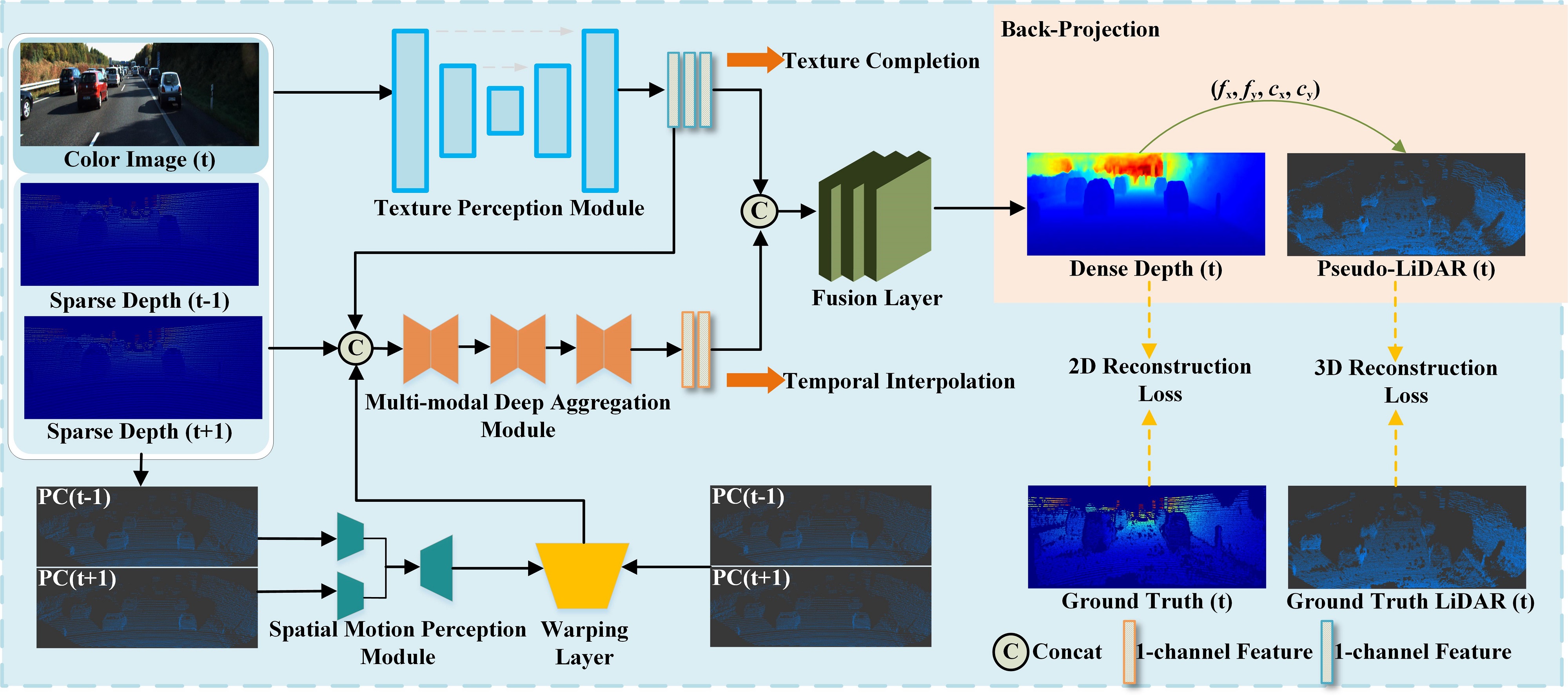}
	\caption{The overview of the proposed Pseudo-LiDAR point cloud interpolation network. Taking an intermediate image at time $t$ and two adjacent sparse depth maps at times $t-1$ and $t+1$ as inputs, the texture completion branch produces the rich texture feature maps. One of the feature maps is used to combine with the sparse depth maps and the scene flow estimated by spatial perception module, to produce the feature maps containing spatial motion information. Finally, a fusion layer fuses feature maps of the dual branch to generate an intermediate dense depth map, which is transformed by back-projecting to provide  our Pseudo-LiDAR point cloud.}
	\label{f2}
\end{figure*}
The remainder of this paper is organized as follows.
In Sec. \ref{sec2}, we describe the related work of point cloud interpolation. In Sec. \ref{sec3}, we introduce the overall model structure and describe each module in detail. Finally, we conduct experiments, and the performance and the results are presented in Sec. \ref{sec4}.

\section{Related Work}\label{sec2}

In this section, related works on the topic of depth estimation, depth completion, and video interpolation will be discussed.

\subsection{Depth Estimation}

Depth estimation focuses on obtaining the depth information of each pixel using a single color image. Earlier works used traditional methods \cite{20liu2014discrete,14karsch2014depth}, which applied hand-crafted features to the depth of color images in probability map models. Recently, with the popularity of convolutional neural networks in image segmentation and detection, learning-based methods have been applied to the depth estimation task. For example, for supervised methods, Eigen et al. \cite{13eigen2015predicting} adopted a multi-scale convolutional architecture to obtain depth information. Li et al. \cite{33li2015depth} used the conditional random field (CRF) post-processing refinement step to perform regression on the features, to obtain high-quality depth output. For unsupervised methods, the supervision is provided by perspective synthesis. Xie et al. \cite{59xie2016deep3d} constructed a stereo pair to calculate and estimate an intermediate disparity image by generating corresponding perspectives. To further improve the performance, \cite{20godard2017unsupervised} used geometric constraints to constrain the consistency of the differences between the left and right images. However, due to the inherent ambiguity and uncertainty of the depth information obtained from color images, the depth map obtained by these depth estimation methods are still inaccurate for navigation systems.

\subsection{Depth Completion}
Compared with the depth estimation, the depth completion task aims to obtain an accurate dense depth map by using a sparse depth map and possible image data. Depth completion covers a series of issues related to different input modalities. When the input modality is relatively dense depth maps that contain missing values, depth completion can be cast as a variety of techniques, such as the executive-based depth inpainting \cite{atapour2018extended, kulkarni2013depth}, object-aware interpolation \cite{atapour2017depthcomp}, Fourier-based depth filling \cite{atapour2016back}, and depth enhancement \cite{camplani2012efficient}.

LiDAR is an indispensable sensor in 3D sensing systems such as autonomous driving and robots. When the acquired LiDAR depth data is projected into the 2D image space, the available depth information only takes up about $4\%$ of the image pixels \cite{Sparsitycnn}. To improve the application of such sparse depth measurements, various methods try to use sparse depth values to estimate dense and accurate depth maps. For example, Uhrig et al. \cite{Sparsitycnn} proposed a simple and effective sparse convolutional layer to take data sparsity into account. Ma et al. \cite{mal2018sparse} considered the depth completion as a regression problem and fed the color image and sparse depth map into an encoder-decoder structure. A similar method in \cite{ma2019self} used a self-supervised training mechanism to achieve the completion task. To extend the confidence of the convolution operation on the continuous network layer, \cite{eldesokey2018propagating} proposed a constrained normalized convolution operating. \cite{huang2019indoor} proposed boundary constraints to enhance the structure and quality of the depth map. \cite{jaritz2018sparse} jointly learned semantic segmentation and completion tasks to improve the performance. In \cite{qiu2019deeplidar}, the surface normal estimation is used for the depth completion task. Chen et al. \cite{chen2019learning} designed an effective network fusion block, which can jointly learn 2D and 3D representations. Compared with these spatial depth completion methods, our method can generate the temporally and spatially high-quality point cloud sequences.

\subsection{Video Interpolation}

In the field of video processing, video interpolation is a popular research topic. Video frame interpolation aims to synthesize the non-existent frames from the original adjacent frames. It makes sense to generate high-quality slow-motion videos from existing videos. Liu et al. \cite{liu2017video} proposed a deep voxel flow network to synthesize video frames by flowing the pixel values of existing frames. To achieve the real-time temporal interpolation, Peleg et al. \cite{5IMNet} adopted an economic structured framework and regarded the interpolation task as a classification problem rather than a regression problem. Jiang et al. \cite{7slomo} jointly modeled the motion interpretation and occlusion inference to achieve variable-length multi-frame video interpolation. Bao et al. \cite{6Depthaware} proposed a depth-aware stream projection layer to guide the detection of occlusion using depth information in the video frame interpolation task. Although there are many works studied in the video frame interpolation, the point cloud sequence interpolation task gains little attention due to the huge volume and complicated structure of point cloud. 

PLIN \cite{liu2020plin} is the first work to interpolate the intermediate Pseudo-LiDAR point cloud given two consecutive stereo pairs. It utilized a coarse-to-fine network structure to facilitate the perception of multi-modal information such as optical flow and color images. Compared with PLIN, our approach uses the improved motion representation, training loss function, and model structure, achieving significant improvements on the Pseudo-LiDAR point cloud interpolation task.

\section{Method}\label{sec3}
In this section, we introduce the overall model structure and describe each module in detail. Given two consecutive sparse depth maps ($D_{t-1}$ and $D_{t+1}$) and RGB image ($I_t$), Pseudo-LiDAR point cloud interpolation aims to produce an intermediate dense depth map $D_t$, which is back-projected to obtain the intermediate Pseudo-LiDAR point cloud ($PC_t$) using known camera parameters. As illustrated in Fig. \ref{f2}, the whole framework mainly consists of two branches: the texture completion branch and temporal interpolation branch. The texture completion branch takes the image and consecutive sparse depth maps as inputs and outputs the feature maps encoded with rich textures. One of the feature maps is further combined with the sparse depth maps and scene flow estimated by spatial perception module, generating the feature maps guided by spatial motion information. The feature maps derived from the dual branch are then integrated by the fusion layer to produce the intermediate dense depth map. Finally, the intermediate Pseudo-LiDAR point cloud is generated by back-projecting the intermediate dense depth map.

\subsection{Texture Completion Branch}
The sparse depth map is difficult to represent the detailed relationship of context information due to its lots of missing pixel values. Therefore, the rich texture information of color images is conducive to the corresponding prediction depth, especially in boundary regions. There are many works that use color images to estimate depth information, which indicates that it can provide corresponding depth inference clues. In our work, to extract the texture and semantic features, the adjacent sparse depth maps and color images are concatenated and fed into the texture completion branch. Moreover, the texture completion branch can be used as a prior to guide the temporal interpolation branch.

We consider the interpolation of Pseudo-LiDAR point cloud as a regression problem. The texture completion network implements an encoder-decoder structure with skip connections. We concatenate the color image and adjacent sparse depth maps into a tensor, which is fed into the residual block of the encoder. In the encoder, the backbone network uses the residual network ResNet-34 \cite{resnet}. In the decoder, the low-dimensional feature maps are up-sampled to the same size as the original feature map through five deconvolution operations. In addition, the multiple skip connections are used to combine low-level features with high-level features. Except for the last layer of convolution, ReLU and BatchNormalization are performed after all convolutions. Finally, the last layer of the texture completion branch uses a $1\times1$ convolution kernel to reduce the multi-channel feature map into a 3-channel feature map. Note that these features only contain the texture and structure information, which cannot describe the accurate motion information yet. Thus, we introduce a spatial motion perception module to further improve the interpolation performance.

\subsection{Spatial Motion Perception Module}
\begin{figure}
	\centering
	\includegraphics[width=1\linewidth]{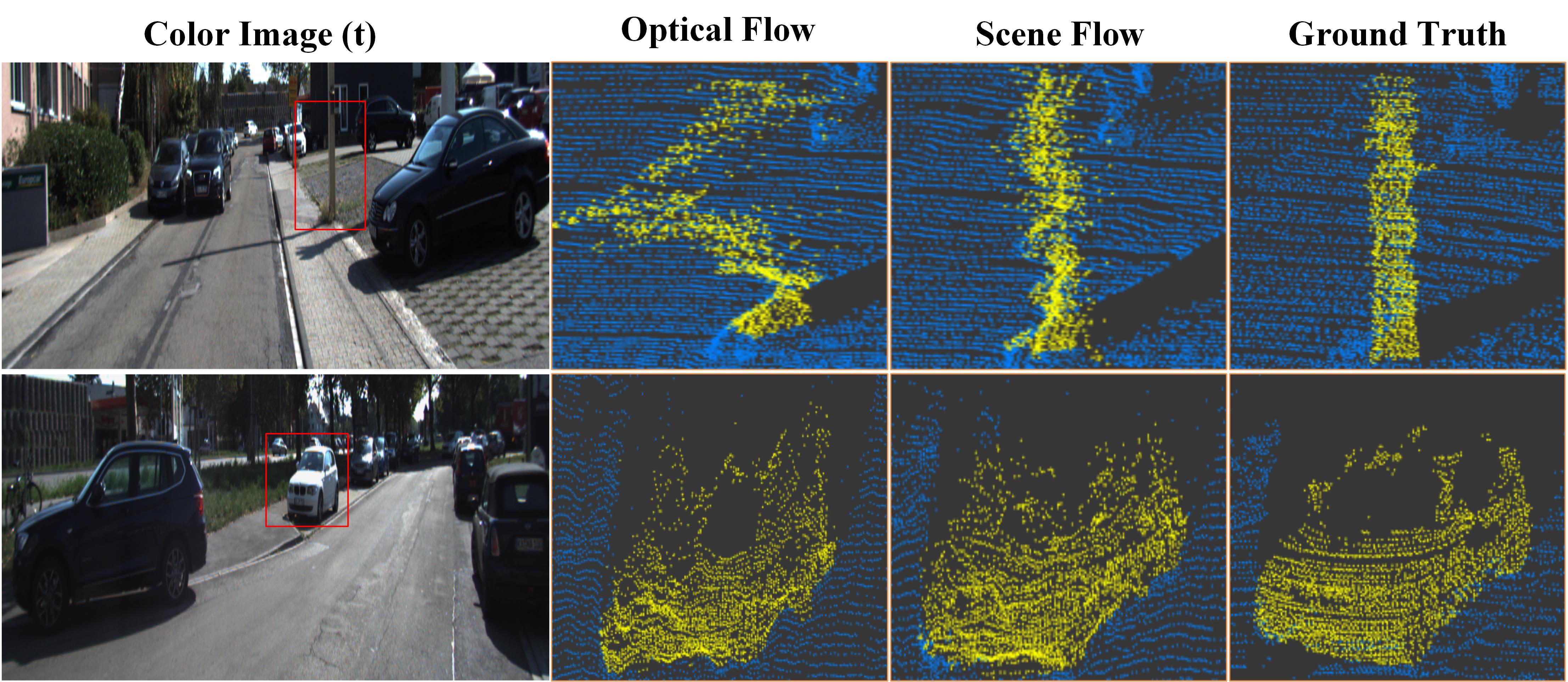}
	\caption{The comparison of Pseudo-LiDAR point cloud guided by different motion information. The result displays the Pseudo-LiDAR point cloud guided by scene flow is more similar to the ground truth, and the distribution is more reasonable.}
	\label{f_flow}
\end{figure}

In the video interpolation task, the optical flow is indispensable since it contains the motion relationship between adjacent frames. Optical flow represents the motion deviation $(\Delta x, \Delta y)$ of the 2D image plane, while the scene flow is represented by the motion field $(\Delta x, \Delta y, \Delta z)$ in 3D space. Scene flow is the counterpart of optical flow in three-dimensional space, it is able to more explicitly represent the real spatial motion relationship of objects. In PLIN, the optical flow between color images is used to represent the motion relationship of depth maps, but the optical flow only represents the deviation of the movement of plane pixels and cannot fully describe the motion information in real 3D space. Therefore, our approach exploits the scene flow to generate a more realistic Pseudo-LiDAR point cloud. As shown in  Fig. \ref{f_flow}, we conducted a comparative experiment. With the same network structure, we use optical flow and scene flow to separately guide the interpolation of the Pseudo-LiDAR point cloud. The results show that the scene flow facilitates to generate a more realistic point cloud. Compared to optical flow, the point cloud guided by the scene flow has a more reasonable shape. This is attributed to better motion representation.

The scene flow estimation method can be described as follows. The input is adjacent point clouds: $PC_{t-1}$ at time $t-1$ and $PC_{t+1}$ at time $t + 1$. Point cloud is a set of points ${(x_i, y_i, z_i)}^n_{i = 1}$, where $n$ is the number of points, and each point may also contain its own attribute features $\left(x_{i}, y_{i}, z_{i}, \ldots\right) \in \mathbb{R}^{d_{f}}$, where $d_f$ refers to the dimension of the attribute feature, such as the reflection intensity, color, and normal. The output is the estimated scene flow $sf_{i}=\left(d x_{i}, d y_{i}, d z_{i}\right)$ for each point $i$ in $PC_{t-1}$. FlowNet3D \cite{liu2019flownet3d} explores the motion based on PointNet++ \cite{Qi2017PointNetDH}, it processes each point and aggregates information through the pooling layers. Our scene flow estimation network is based on FlowNet3D and the improved bilateral convolutional layer operations, which restore the spatial information from unstructured point clouds. The input of the network is 3D point clouds of consecutive frames, and the output is the corresponding deviation of each point. Scene flow calculations can be expressed as follows.

\begin{equation}
\begin{aligned} 
	sf_{t-1\rightarrow t+1} = H^{sf}(PC_{t-1},PC_{t+1}),\\
	sf_{t+1\rightarrow t-1} = H^{sf}(PC_{t+1},PC_{t-1}),
\end{aligned} 
\end{equation}
where $PC_{t-1}$ and $PC_{t+1}$ denote the input point clouds of adjacent frames, $H^{sf}$ is the scene flow estimation function, $sf_{t-1\rightarrow t+1}, sf_{t+1\rightarrow t-1}$ refer to the estimated scene flow.

Since the input of the scene flow estimation task is the adjacent point clouds, we first convert the adjacent depth maps into point clouds in terms of the prior camera parameters. We will discuss the specific transformation formulas in Section \ref{transformation}. The adjacent point clouds are then inputted to our spatial motion perception module to estimate the scene flow. Our scene flow estimation network is similar to the encoder-decoder structure. In the downsampling stage, we adopt a dual-input structure, in which all layers share weights to extract the features of point clouds. By stacking improved bilateral convolutional layer (BCL)\cite{kiefel2015permutohedral} to continuously reduce the scale. We also fuse features of different scales. In the upsampling phase, we gradually increase the scale by stacking the improved bilateral convolutional layer to improve the prediction. In each BCL, we consider the relative position of the input. Finally, our scene flow is obtained. We use a warping operation on $PC_{t-1}$ or $PC_{t+1}$ to synthesize the point cloud $PC_{t}$ at time $t$, which can be expressed as:
\begin{equation}
PC_{t} =PC_{t-1}+ \frac{sf_{t-1\rightarrow t+1}}{2},\\
\end{equation}
\centerline{or}
\begin{equation}
PC_{t} =PC_{t+1}+ \frac{sf_{t-1\rightarrow t+1}}{2}.\\
\end{equation}

To boost the fast spatial information sensing, we project the obtained intermediate point cloud $PC_t$ into the 2D image plane. In this part, we get the accurate but sparse intermediate depth map $D_t$. To effectively integrate multi-modal features and generate an accurate and dense depth map, we introduce a multi-modal deep aggregation module to facilitate the efficient fusion of texture and depth features.

\subsection{Multi-modal Deep Aggregation Module}
To generate the accurate and dense depth map, we design a multi-modal deep aggregation module to fuse the feature maps of the texture completion branch and the temporal interpolation branch. The texture feature can guide the network to pay more attention to the saliency objects, which contains the more clear structure and edge information. On the other hand, the depth feature can provide precise spatial information in terms of the estimated scene flow.

In particular, we adopt a stacked aggregation unit architecture for the multi-modal deep aggregation module. The stacked aggregation unit consists of three aggregation units, each of which has a top-down and bottom-up process. Inspired by ResNet \cite{resnet}, we use a residual learning method between aggregation units. In addition, the skip connection operations are applied to introduce the low-level feature into the high-level feature in the same dimension. 
 
In each aggregation unit, the encoder and decoder are composed of three layers of convolutions. The encoder uses two stride convolutions to downsample the feature resolution to the 1/4 original size; the decoder uses two deconvolution operations to upsample the features fused from the encoder and the previous network block. Considering the sparseness of data, the encoder in the first network block does not use the batch normalization operation after convolution. All convolutional layers use a 3$\times$3 convolution kernel with a small receptive field. The output of the multi-modal deep aggregation module is a 2-channel feature map containing dense spatial distribution information.

At the end of the dual branch architecture, we leverage a fusion layer to further combine the different feature maps and obtain the final result. The fusion layer consists of three convolutional layers and the number of filters per convolutional layer is 32, 32, and 1, respectively. Except for the last layer, the BatchNormalization layer with ReLU activation function is implemented after each convolutional layer. 

\begin{figure*}[t]
	\centering
	\includegraphics[width=1\linewidth]{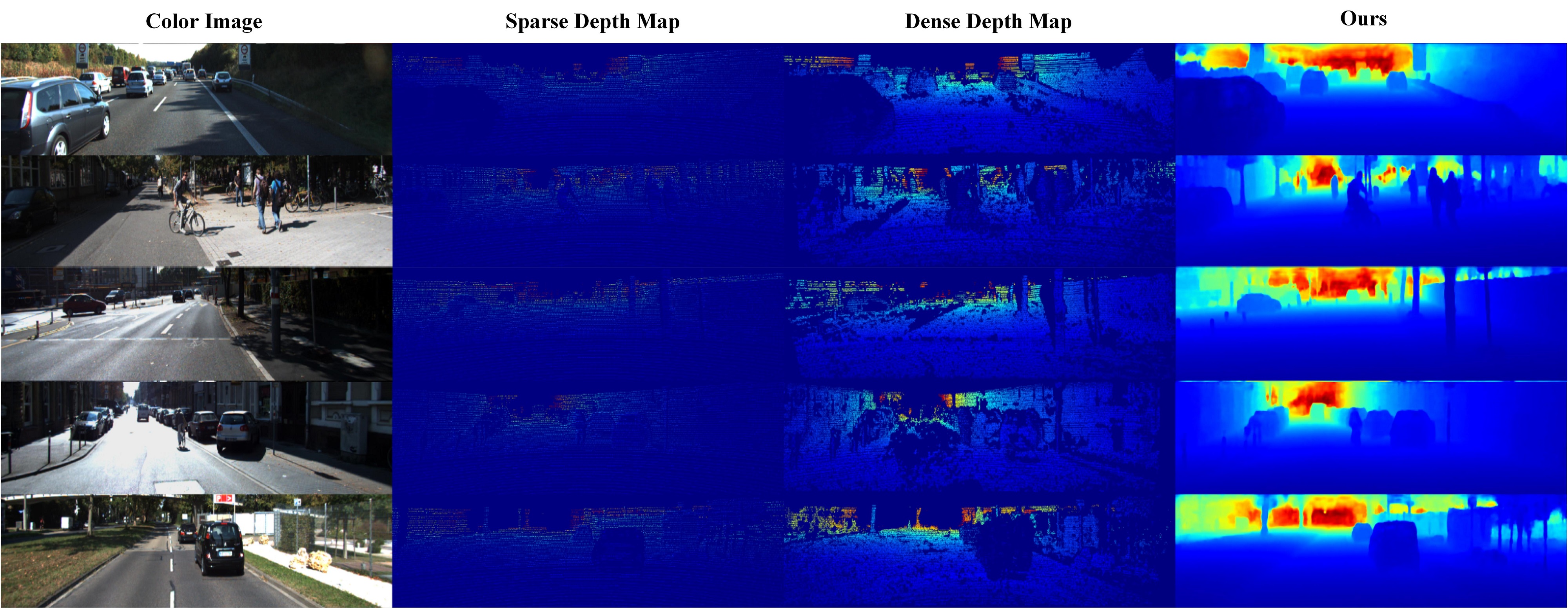}
	\caption{Results of the interpolated depth map obtained by our approach. For each example, we show the color image, sparse depth map, dense depth map, and our interpolated result. The dense depth map represents the ground truth of training. Our depth map results have denser distributions and clear boundaries of objects.}
	\label{f3}
\end{figure*}
\subsection{Back-Projection} \label{transformation}

In this part, we get the 3D point cloud by back-projecting the generated intermediate depth map to 3D space. According to the pinhole camera imaging principle, if the depth value $Z_{t}(u, v)$ of each pixel coordinate $(u, v)$ exists in the image, we can derive the corresponding 3D position $(x, y, z)$. The corresponding relationship is described as follows:
\begin{align} 
	x &=\frac{\left(u-c_{u}\right) \times z}{f_{u}}, \\ 
	y &=\frac{\left(v-c_{v}\right) \times z}{f_{v}}, \\
	z &=Z_t(u, v),
\end{align}
where $f_v$ and $f_u$ are the vertical and horizontal focal lengths, respectively. $(c_u, c_v)$ is the center of camera aperture. Based on the prior camera parameters, the generated depth map is back-projected into a 3D point cloud. Since this point cloud is obtained by transforming the depth map, we refer to the point cloud as a Pseudo-LiDAR point cloud \cite{wang2019pseudo}.

\subsection{Loss Function}
Previous work only supervises the generated dense depth maps, which does not constrain the 3D structure of the target point cloud. To this end, we design a point cloud reconstruction loss to supervise the generation of Pseudo-LiDAR point clouds. Constructing the distance function between the predicted point cloud and the ground truth point cloud is an important step. A suitable distance function should meet at least two conditions: 1) the calculation is differentiable; 2) since data needs to be forwarded and back-propagated for many times, effective calculations are required \cite{fan2017point}. The goal of our efforts can be expressed as:

\begin{equation}
L\left(\left\{PC_{i}^{p r e d}\right\},\left\{PC_{i}^{g t}\right\}\right)=\sum d\left(PC_{i}^{p r e d}, PC_{i}^{g t}\right),
\end{equation}
where $PC_{i}^{p r e d}$ and $PC_{i}^{g t}$ indicate the prediction and ground truth of each sample, respectively.

We need to find a distance metric $d \subseteq \mathbb{R}^{3}$ to minimize the difference between the generated point cloud and the ground truth point cloud. There are two candidates for the measurement: Earth Mover’s distance (EMD) and Chamfer distance (CD):

Earth Mover’s distance: if two point sets $PC_{1}, PC_{2} \subseteq \mathbb{R}^{3}$ and have the same size. The EMD can be defined as:
\begin{equation}
d_{E M D}\left(PC_{1}, PC_{2}\right)=\min _{\phi: PC_{1} \rightarrow PC_{2}} \sum_{x \in PC_{1}}\|x-\phi(x)\|_{2},
\end{equation}
where $\phi: PC_{1} \rightarrow PC_{2}$ is a bijection. EMD is almost differentiable everywhere, but its accurate calculation is expensive for learning models.

Chamfer distance: we can define it between $PC_{1}, PC_{2} \subseteq \mathbb{R}^{3}$ as:
\begin{equation}
\begin{aligned}
d_{C D}\left(PC_{1}, PC_{2}\right)=\sum_{x \in PC_{1}} \min _{y \in PC_{2}}\|x-y\|_{2}^{2}+\\
\sum_{y \in PC_{2}} \min _{x \in PC_{1}}\|x-y\|_{2}^{2}.
\end{aligned}
\end{equation}

\noindent This algorithm finds the nearest point of each point $PC_1$ in another point set $PC_2$ and adds up the squared distances. For each point, searching for the nearest point is independent and easily parallelized. To speed up the nearest point search, a similar KD-tree data structure can be applied. Since EDM has a limitation on the number of input points, we use the simple and effective CD distance as our reconstruction loss to evaluate the similarity between generated point cloud and ground truth point cloud. Our reconstruction loss is formed as follows.
\begin{equation}
\begin{aligned}
\mathcal{L}_{rec}(PC_{pred}, PC_{gt}) = 
  &d_{C D}\left(PC_{pred}, PC_{gt}\right)\\
+ &d_{C D}\left(PC_{gt},PC_{pred}\right),
\end{aligned}
\end{equation}
where $d_{C D}$ denotes the chamfer distance metric. $PC_{p r e d}$ and $PC_{g t}$ are the prediction and ground truth point cloud, respectively.

In addition to the point cloud supervision, we also perform the 2D supervision on dense depth maps. We use $L_2$ loss for the generated depth map $d_{pred}$ and ground truth $d_{gt}$:
\begin{equation}
\mathcal{L}_{d}(d_{pred}, d_{gt})=\| 1_{\{d_{gt}>0\}} \cdot( d_{pred}-d_{gt}) \|_{2}^{2}.
\end{equation}

Our entire loss function is a linear combination of point cloud reconstruction loss and depth map reconstruction loss, which can be expressed as:
\begin{equation}
\mathcal{L} = w_1\cdot\mathcal{L}_{ d}(d_{pred}, gt)+w_2\cdot \mathcal{L}_{rec}(PC_{pred}, PC_{gt}),
\end{equation}
where $w_1$ and $w_2$ are the balance weights. The weights have been set empirically as $w_1=1$ and $w_2=1$.

\section{Experiments}\label{sec4}
In this section, we conduct extensive experiments to verify the effectiveness of our proposed approach. We compare with previous works and perform a series of ablation studies to show the effectiveness of each module. Since the main application of our model is on-board LiDARs in a multi-sensor system, our experiments are based on the KITTI dataset \cite{kitti}. As illustrated in Fig.\ref{f3}, the depth maps obtained by our approach show clear boundaries in visual effects and display denser distributions than the ground truth dense depth maps.
\begin{figure*}[t]
	\centering
	\includegraphics[width=1\linewidth]{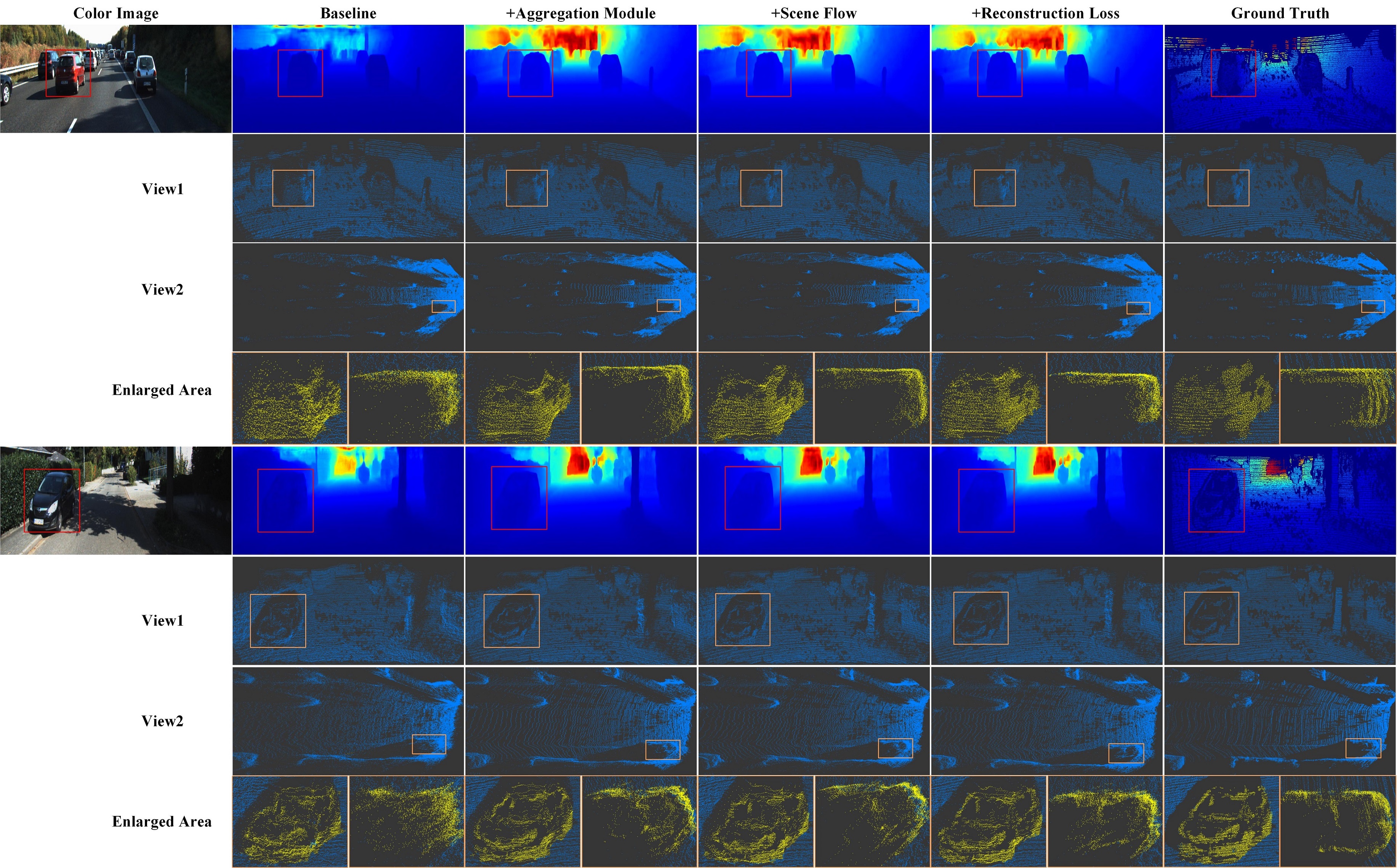}
	\caption{Visual results of the ablation study. For each configuration, from top to bottom are depth maps corresponding to color images, point cloud view 1, point cloud view 2, and partially zoomed regions. Our method produces a reasonable distribution and shape of Pseudo-LiDAR point clouds.}
	\label{f4}
\end{figure*}
\subsection{Experimental Setting}
\noindent{\textbf{Dataset:}} Our experiments are performed on the KITTI depth completion dataset and the raw dataset. The KITTI dataset contains 86,898 frames of training data, 6,852 frames of evaluation data, and 1,000 frames of test data. This dataset provides sparse depth maps and color images. Each frame contains LiDAR scan data and RGB color images, in which the sparse depth map corresponds to the projection of the 3D LiDAR scan point cloud. The ground truth corresponding to each sparse depth map is a relatively dense depth map. Our application scenario is based on the outdoor on-board LiDAR, which is generally a scene of relative motion. Since there are scenes where the frames are still in the training dataset, so we choose 48,000 frames with obvious motions.

\noindent{\textbf{Evaluation Metrics:}} Although our task is not the depth completion, we can use the evaluation metrics of depth completion to evaluate the quality of the generated dense depth map. There are four evaluation metrics in the depth completion task: the root mean square error (RMSE), mean absolute error (MAE), root mean square error inverse depth (iRMSE), and mean absolute error inverse depth (iMAE) . We mainly focus on the RMSE when comparing methods because RMSE directly measures the error in depth, penalizes larger errors, and is the leading metric for depth completion. These four evaluation indicators are defined by the following formulas:

\begin{itemize}
\item  Root mean squared error (RMSE):

\begin{equation}
R M S E=\sqrt{\frac{1}{n} \sum\left(d_{p r e d}-d_{g t}\right)^{2}}
\end{equation}
\item  Mean absolute error (MAE):

\begin{equation}
	MAE=\frac{1}{n} \sum\left|d_{pred}-d_{gt}\right|
\end{equation}
\item  Root mean squared error of the inverse depth [1/km](iRMSE):

\begin{equation}
i R M S E=\sqrt{\frac{1}{n} \sum\left(\frac{1}{d_{ pred }}-\frac{1}{d_{g t}}\right)^{2}}
\end{equation}
\item  Mean absolute error of the inverse depth [1/km](iMAE):

\begin{equation}
i M A E=\frac{1}{n} \sum\left|\frac{1}{d_{p r e d}}-\frac{1}{d_{g t}}\right|
\end{equation}
\end{itemize}

In order to evaluate the quality of the generated point cloud, we introduce a new evaluation metrics, i.e., CD as follows:

\begin{equation}
CD=d_{C D}\left(PC_{1}, PC_{2}\right)+d_{C D}\left(PC_{2}, PC_{1}\right)
\end{equation}

\begin{figure*}[t]
	\centering
	\includegraphics[width=1\linewidth]{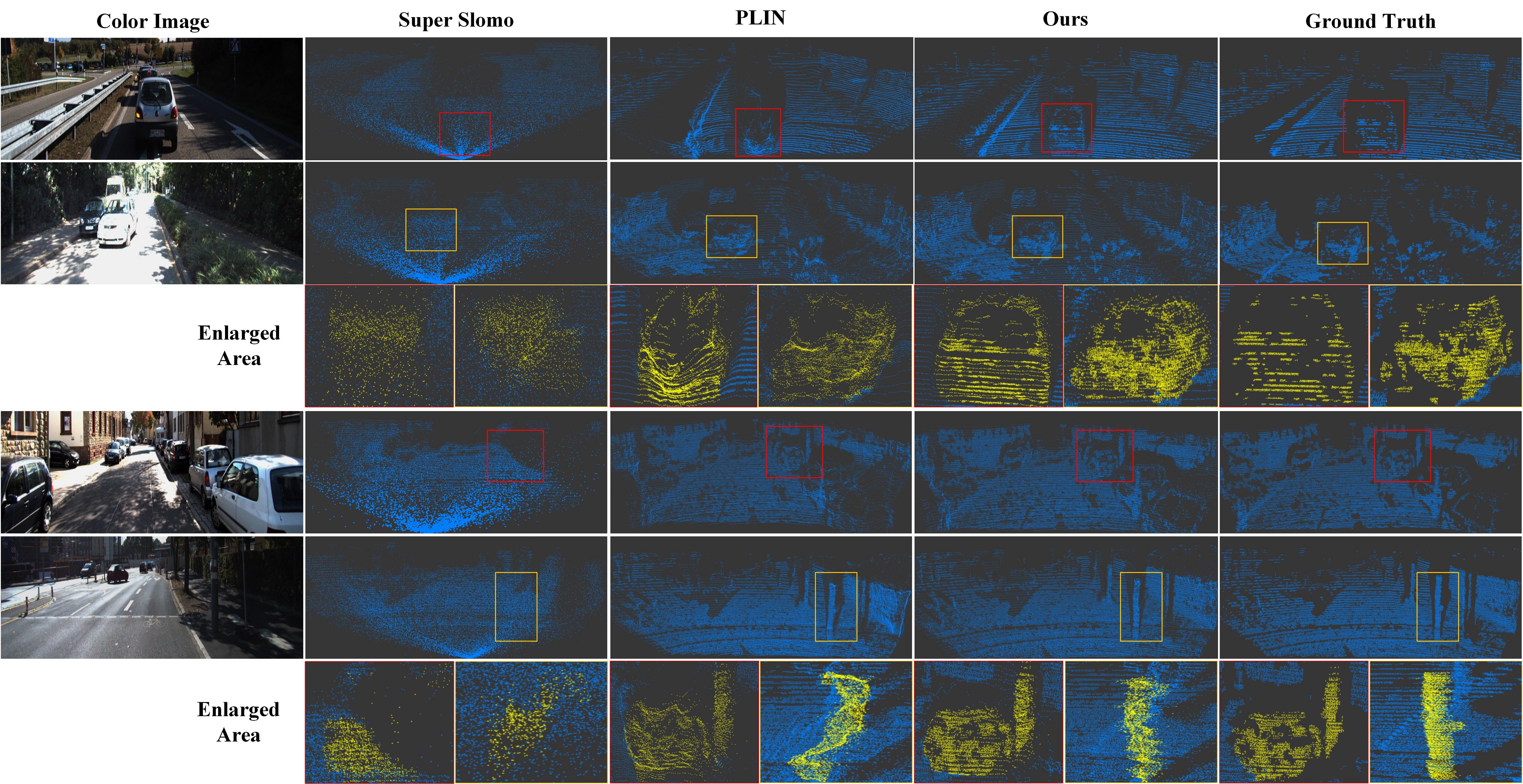}
	\caption{Visual comparison with state-of-the-art methods (better viewed in color). For each scene, we show the color image and the Pseudo-LiDAR point cloud obtained by different methods. In the third row, we enlarged the local regions for better observation. Our method produces a more reasonable distribution and shape. In the zoomed regions, our method recovers better 3D details. }
	\label{f5}
\end{figure*}

\noindent{\textbf{Implementation Details:}} The depth value at the upper end of the depth map is all zero, and this section does not provide any depth information. Therefore, all our data (RGB, sparse depth, and dense depth map) are cropped from the bottom to a uniform 1216$\times$256 size. Data enhancement operations are also applied to the training data, such as random flips and color adjustments. In the calculation of scene flow, we randomly sample 17,500 points in the point cloud of each frame as the input of the scene flow network, which is designed based on the HPLFlowNet \cite{gu2019hplflownet}. Adam optimizer is applied during our training phase with $10^{-4}$ initial learning rate, which is decayed by 0.1 every 4 epochs. We train our network on a 1080Ti GPU with a batch size of 2 for about 60 hours, which is completed by PyTorch \cite{paszke2017automatic}.

\begin{table}[htbp]
	\caption{Ablation study: performance achieved by our network  with and without each module.}
	\label{table_1}
	\begin{center}
		\scalebox{0.96}{
		\begin{tabular}{c|c|c|c|c|c}
			\hline
			Configuration & RMSE & MAE & iRMSE &iMAE&CD\\
			\hline
			Baseline & 1408.80 & 513.06 & 7.63 & 3.01&0.21\\
		   +Aggregation module & 1224.91 & 409.69 & 4.69 & 1.95&0.16 \\
			+Scene flow & 1124.76 & 382.15 & 4.39 & 1.89&0.14\\
			+Reconstruction loss &\textbf{1091.99} & \textbf{371.56} & \textbf{4.21} & \textbf{1.83}&\textbf{0.12}\\
			\hline
		\end{tabular}}
	\end{center}
\end{table}

\subsection{Ablation Study}
We perform an extensive ablation study to verify the effectiveness of each module. The performance comparison of the proposed approach is shown in Table \ref{table_1}. Specifically, we perform four ablation experiments, each of which is based on the addition of a new network element or module to the previous network configuration.

As listed in Table \ref{table_1}, the result shows that the complete network achieves the best interpolation performance. For the baseline network, we take two consecutive sparse depth maps and the intermediate color image as the input of the texture completion branch and obtain the intermediate dense depth map. By comparing the experimental results, we have the following observations: 1) Without the spatial motion guidance, our multi-modal deep aggregation module can also produce good interpolation results, as it combines the features of the dual branch and is more conducive to the fusion of features. 2) Under the guidance of the scene flow containing motion information, we have greatly improved the performance of interpolation. This benefits from a better representation of spatial motion information. 3) Point cloud reconstruction constraints also further improve the interpolation performance. It can be observed that the value of our evaluation metrics decreases as the number of modules increases, which also proves the effectiveness of each of our network modules. To intuitively compare these different performances, we visualize the interpolated results of two scenes obtained by the above methods in Fig. \ref{f4}. The complete network generates the most realistic details and distributions of the intermediate point cloud. Note that in the enlarged area, the shape distribution of the car obtained by the complete network is the most similar to the ground truth.

\begin{table}[htbp]
	\caption{Quantative evaluation results of the traditional interpolation method, Super Slomo \cite{7slomo},PLIN \cite{liu2020plin}, and our method.}
	\label{table_2}
	\begin{center}
		\scalebox{0.86}{
		\begin{tabular}{c|c|c|c|c|c}
			\hline
			\textbf{Method}&RMSE& MAE &iRMSE & iMAE& CD\\
			\hline
			Traditional Interpolation & 12552.46 & 3868.80 & - & -&-\\
			Super Slomo \cite{7slomo} & 16055.19 & 11692.72 & - & - &27.38\\
			PLIN\cite{liu2020plin} &1168.27 & 546.37 & 6.84 & 3.68 &0.21\\
			Ours &\textbf{1091.99} & \textbf{371.56} & \textbf{4.21} & \textbf{1.83}& \textbf{0.12}\\
			\hline
		\end{tabular} }
	\end{center}
\end{table}

\subsection{Comparison with State-of-the-art}
We evaluate our model on the KITTI depth completion dataset. We show the comparison results with other state-of-the-art point cloud interpolation methods in Table \ref{table_2}. Since PLIN is a pioneer work in this field, it is our main comparison object. In addition, we also compare the traditional average depth interpolation method and video interpolation method. For the video frame interpolation method, the Super Slomo \cite{7slomo} network is retrained on the depth completion dataset.

\noindent \textbf{Quantitative Comparison.}
We show some quantitative results comparing our proposed approach with existing methods in Table \ref{table_2}. Experimental results show that our approach is superior to other methods in learning the interpolation of the point cloud from RGBD data. In particular, we achieve state-of-the-art results in all metrics. For the traditional method, the intermediate depth map is obtained by averaging consecutive depth maps. Its poor performance is understandable because the pixel values of continuous depth maps do not have a corresponding relationship. For the video frame interpolation method, since the motion relationship between depth maps cannot be obtained, it is difficult to generate satisfactory results. Guided by the color images and bidirectional optical flow, PLIN is designed for the task of point cloud interpolation and achieves good performance, but it lacks the point cloud supervision and spatial motion representation. Compared with these methods, our approach improves the Pseudo-LiDAR point cloud interpolation task by adopting the scene flow, 3D space supervision mechanism, and multi-modal deep aggregation module. As a result, our approach outperforms the classical methods.

\noindent \textbf{Visual Comparison.}
For the visual comparison, we compare different interpolation results in Fig. \ref{f5}. In PLIN, it has been shown that the traditional interpolation method cannot handle the point cloud interpolation problem well, and the visual performance is poor. Therefore, we only show the comparison on Super Slomo \cite{7slomo}, PLIN, our approach, and ground truth. As illustrated in Fig. \ref{f5}, our approach produces a more reasonable distribution and shape compared with PLIN. The whole distribution of Pseudo-LiDAR point cloud is more similar to that of the ground truth point cloud. In the zoomed regions, our method recovers better 3D details for car, road, and tree. This benefited from optimized motion representation, 3D space supervision mechanism and model structure.

\section{Conclusion}
In this paper, we propose a novel Pseudo-LiDAR point cloud interpolation network with better interpolation performance than previous works. To more accurately represent the spatial motion information, we use the point cloud scene flow to guide the point cloud interpolation task. We design a multi-modal deep aggregation module to facilitate the efficient fusion of features of the dual branch. In addition, we adopt a supervision mechanism in 3D space to supervise the generation of Pseudo-LiDAR point cloud. As the benefits of the optimized motion representation, training loss function, and model structure, the proposed pipeline significantly improves the performance of interpolation. We have shown the effectiveness of our approach on the KITTI dataset, outperforming the state-of-the-art point cloud interpolation techniques with a large margin.

\appendices
%

%
%

\ifCLASSOPTIONcaptionsoff
  \newpage
\fi



%
%

\bibliographystyle{IEEEtran}
\bibliography{reference}

\begin{thebibliography}{10}
\providecommand{\url}[1]{#1}
\csname url@samestyle\endcsname
\providecommand{\newblock}{\relax}
\providecommand{\bibinfo}[2]{#2}
\providecommand{\BIBentrySTDinterwordspacing}{\spaceskip=0pt\relax}
\providecommand{\BIBentryALTinterwordstretchfactor}{4}
\providecommand{\BIBentryALTinterwordspacing}{\spaceskip=\fontdimen2\font plus
\BIBentryALTinterwordstretchfactor\fontdimen3\font minus
  \fontdimen4\font\relax}
\providecommand{\BIBforeignlanguage}[2]{{%
\expandafter\ifx\csname l@#1\endcsname\relax
\typeout{** WARNING: IEEEtran.bst: No hyphenation pattern has been}%
\typeout{** loaded for the language `#1'. Using the pattern for}%
\typeout{** the default language instead.}%
\else
\language=\csname l@#1\endcsname
\fi
#2}}
\providecommand{\BIBdecl}{\relax}
\BIBdecl

\bibitem{yang2019reactive}
X.~Yang, H.~Luo, Y.~Wu, Y.~Gao, C.~Liao, and K.-T. Cheng, ``Reactive obstacle
  avoidance of monocular quadrotors with online adapted depth prediction
  network,'' \emph{Neurocomputing}, vol. 325, pp. 142--158, 2019.

\bibitem{Pointrcnn}
S.~Shi, X.~Wang, and H.~Li, ``Point{R}cnn: 3{D} object proposal generation and
  detection from point cloud,'' in \emph{Proceedings of the IEEE Conference on
  Computer Vision and Pattern Recognition (CVPR)}, 2019, pp. 770--779.

\bibitem{fpointnet}
C.~R. Qi, W.~Liu, C.~Wu, H.~Su, and L.~J. Guibas, ``Frustum pointnets for 3{D}
  object detection from {RGB-D} data,'' in \emph{Proceedings of the IEEE
  Conference on Computer Vision and Pattern Recognition (CVPR)}, 2018, pp.
  918--927.

\bibitem{shin20193d}
D.~Shin, Z.~Ren, E.~B. Sudderth, and C.~C. Fowlkes, ``3d scene reconstruction
  with multi-layer depth and epipolar transformers,'' in \emph{Proceedings of
  the IEEE International Conference on Computer Vision}, 2019, pp. 2172--2182.

\bibitem{liu2020plin}
H.~Liu, K.~Liao, C.~Lin, Y.~Zhao, and M.~Liu, ``Plin: A network for
  pseudo-lidar point cloud interpolation,'' \emph{Sensors}, vol.~20, no.~6, p.
  1573, 2020.

\bibitem{20liu2014discrete}
M.~Liu, M.~Salzmann, and X.~He, ``Discrete-continuous depth estimation from a
  single image,'' in \emph{Proceedings of the IEEE Conference on Computer
  Vision and Pattern Recognition}, 2014, pp. 716--723.

\bibitem{14karsch2014depth}
K.~Karsch, C.~Liu, and S.~B. Kang, ``Depth transfer: Depth extraction from
  video using non-parametric sampling,'' \emph{IEEE transactions on pattern
  analysis and machine intelligence}, vol.~36, no.~11, pp. 2144--2158, 2014.

\bibitem{13eigen2015predicting}
D.~Eigen and R.~Fergus, ``Predicting depth, surface normals and semantic labels
  with a common multi-scale convolutional architecture,'' in \emph{Proceedings
  of the IEEE international conference on computer vision}, 2015, pp.
  2650--2658.

\bibitem{33li2015depth}
B.~Li, C.~Shen, Y.~Dai, A.~Van Den~Hengel, and M.~He, ``Depth and surface
  normal estimation from monocular images using regression on deep features and
  hierarchical crfs,'' in \emph{Proceedings of the IEEE conference on computer
  vision and pattern recognition}, 2015, pp. 1119--1127.

\bibitem{59xie2016deep3d}
J.~Xie, R.~Girshick, and A.~Farhadi, ``Deep3d: Fully automatic 2d-to-3d video
  conversion with deep convolutional neural networks,'' in \emph{European
  Conference on Computer Vision}.\hskip 1em plus 0.5em minus 0.4em\relax
  Springer, 2016, pp. 842--857.

\bibitem{20godard2017unsupervised}
C.~Godard, O.~Mac~Aodha, and G.~J. Brostow, ``Unsupervised monocular depth
  estimation with left-right consistency,'' in \emph{Proceedings of the IEEE
  Conference on Computer Vision and Pattern Recognition}, 2017, pp. 270--279.

\bibitem{atapour2018extended}
A.~Atapour-Abarghouei and T.~P. Breckon, ``Extended patch prioritization for
  depth filling within constrained exemplar-based rgb-d image completion,'' in
  \emph{International Conference Image Analysis and Recognition}.\hskip 1em
  plus 0.5em minus 0.4em\relax Springer, 2018, pp. 306--314.

\bibitem{kulkarni2013depth}
M.~Kulkarni and A.~N. Rajagopalan, ``Depth inpainting by tensor voting,''
  \emph{JOSA A}, vol.~30, no.~6, pp. 1155--1165, 2013.

\bibitem{atapour2017depthcomp}
A.~Atapour-Abarghouei and T.~P. Breckon, ``Depthcomp: real-time depth image
  completion based on prior semantic scene segmentation.'' 2017.

\bibitem{atapour2016back}
A.~Atapour-Abarghouei, G.~P. de~La~Garanderie, and T.~P. Breckon, ``Back to
  butterworth-a fourier basis for 3d surface relief hole filling within rgb-d
  imagery,'' in \emph{2016 23rd International Conference on Pattern Recognition
  (ICPR)}.\hskip 1em plus 0.5em minus 0.4em\relax IEEE, 2016, pp. 2813--2818.

\bibitem{camplani2012efficient}
M.~Camplani and L.~Salgado, ``Efficient spatio-temporal hole filling strategy
  for kinect depth maps,'' in \emph{Three-dimensional image processing (3DIP)
  and applications Ii}, vol. 8290.\hskip 1em plus 0.5em minus 0.4em\relax
  International Society for Optics and Photonics, 2012, p. 82900E.

\bibitem{Sparsitycnn}
J.~Uhrig, N.~Schneider, L.~Schneider, U.~Franke, T.~Brox, and A.~Geiger,
  ``Sparsity invariant cnns,'' in \emph{2017 International Conference on 3D
  Vision (3DV)}.\hskip 1em plus 0.5em minus 0.4em\relax IEEE, 2017, pp. 11--20.

\bibitem{mal2018sparse}
F.~Mal and S.~Karaman, ``Sparse-to-dense: Depth prediction from sparse depth
  samples and a single image,'' in \emph{2018 IEEE International Conference on
  Robotics and Automation (ICRA)}.\hskip 1em plus 0.5em minus 0.4em\relax IEEE,
  2018, pp. 1--8.

\bibitem{ma2019self}
F.~Ma, G.~V. Cavalheiro, and S.~Karaman, ``Self-supervised sparse-to-dense:
  Self-supervised depth completion from lidar and monocular camera,'' in
  \emph{2019 International Conference on Robotics and Automation (ICRA)}.\hskip
  1em plus 0.5em minus 0.4em\relax IEEE, 2019, pp. 3288--3295.

\bibitem{eldesokey2018propagating}
A.~Eldesokey, M.~Felsberg, and F.~S. Khan, ``Propagating confidences through
  cnns for sparse data regression,'' \emph{arXiv preprint arXiv:1805.11913},
  2018.

\bibitem{huang2019indoor}
Y.-K. Huang, T.-H. Wu, Y.-C. Liu, and W.~H. Hsu, ``Indoor depth completion with
  boundary consistency and self-attention,'' in \emph{Proceedings of the IEEE
  International Conference on Computer Vision Workshops}, 2019, pp. 0--0.

\bibitem{jaritz2018sparse}
M.~Jaritz, R.~De~Charette, E.~Wirbel, X.~Perrotton, and F.~Nashashibi, ``Sparse
  and dense data with cnns: Depth completion and semantic segmentation,'' in
  \emph{2018 International Conference on 3D Vision (3DV)}.\hskip 1em plus 0.5em
  minus 0.4em\relax IEEE, 2018, pp. 52--60.

\bibitem{qiu2019deeplidar}
J.~Qiu, Z.~Cui, Y.~Zhang, X.~Zhang, S.~Liu, B.~Zeng, and M.~Pollefeys,
  ``Deeplidar: Deep surface normal guided depth prediction for outdoor scene
  from sparse lidar data and single color image,'' in \emph{Proceedings of the
  IEEE Conference on Computer Vision and Pattern Recognition}, 2019, pp.
  3313--3322.

\bibitem{chen2019learning}
Y.~Chen, B.~Yang, M.~Liang, and R.~Urtasun, ``Learning joint 2d-3d
  representations for depth completion,'' in \emph{Proceedings of the IEEE
  International Conference on Computer Vision}, 2019, pp. 10\,023--10\,032.

\bibitem{liu2017video}
Z.~Liu, R.~A. Yeh, X.~Tang, Y.~Liu, and A.~Agarwala, ``Video frame synthesis
  using deep voxel flow,'' in \emph{Proceedings of the IEEE International
  Conference on Computer Vision}, 2017, pp. 4463--4471.

\bibitem{5IMNet}
T.~Peleg, P.~Szekely, D.~Sabo, and O.~Sendik, ``Im-net for high resolution
  video frame interpolation,'' in \emph{Proceedings of the IEEE Conference on
  Computer Vision and Pattern Recognition (CVPR)}, 2019, pp. 2398--2407.

\bibitem{7slomo}
H.~Jiang, D.~Sun, V.~Jampani, M.-H. Yang, E.~Learned-Miller, and J.~Kautz,
  ``Super slomo: High quality estimation of multiple intermediate frames for
  video interpolation,'' in \emph{Proceedings of the IEEE Conference on
  Computer Vision and Pattern Recognition (CVPR)}, 2018, pp. 9000--9008.

\bibitem{6Depthaware}
W.~Bao, W.-S. Lai, C.~Ma, X.~Zhang, Z.~Gao, and M.-H. Yang, ``Depth-aware video
  frame interpolation,'' in \emph{Proceedings of the IEEE Conference on
  Computer Vision and Pattern Recognition (CVPR)}, 2019, pp. 3703--3712.

\bibitem{resnet}
K.~He, X.~Zhang, S.~Ren, and J.~Sun, ``Deep residual learning for image
  recognition,'' in \emph{Proceedings of the IEEE Conference on Computer Vision
  and Pattern Recognition (CVPR)}, 2016, pp. 770--778.

\bibitem{liu2019flownet3d}
X.~Liu, C.~R. Qi, and L.~J. Guibas, ``Flownet3d: Learning scene flow in 3d
  point clouds,'' in \emph{Proceedings of the IEEE Conference on Computer
  Vision and Pattern Recognition}, 2019, pp. 529--537.

\bibitem{Qi2017PointNetDH}
C.~R. Qi, L.~Yi, H.~Su, and L.~J. Guibas, ``Pointnet++: Deep hierarchical
  feature learning on point sets in a metric space,'' in \emph{NeurIPS}, 2017.

\bibitem{kiefel2015permutohedral}
M.~Kiefel, V.~Jampani, and P.~V. Gehler, ``Permutohedral lattice cnns,'' 2015.

\bibitem{wang2019pseudo}
Y.~Wang, W.-L. Chao, D.~Garg, B.~Hariharan, M.~Campbell, and K.~Q. Weinberger,
  ``Pseudo-lidar from visual depth estimation: Bridging the gap in 3d object
  detection for autonomous driving,'' in \emph{Proceedings of the IEEE
  Conference on Computer Vision and Pattern Recognition}, 2019, pp. 8445--8453.

\bibitem{fan2017point}
H.~Fan, H.~Su, and L.~J. Guibas, ``A point set generation network for 3d object
  reconstruction from a single image,'' in \emph{Proceedings of the IEEE
  conference on computer vision and pattern recognition}, 2017, pp. 605--613.

\bibitem{kitti}
A.~Geiger, P.~Lenz, C.~Stiller, and R.~Urtasun, ``Vision meets robotics: The
  {KITTI} dataset,'' \emph{The International Journal of Robotics Research},
  vol.~32, no.~11, pp. 1231--1237, 2013.

\bibitem{gu2019hplflownet}
X.~Gu, Y.~Wang, C.~Wu, Y.~J. Lee, and P.~Wang, ``Hplflownet: Hierarchical
  permutohedral lattice flownet for scene flow estimation on large-scale point
  clouds,'' in \emph{Proceedings of the IEEE Conference on Computer Vision and
  Pattern Recognition}, 2019, pp. 3254--3263.

\bibitem{paszke2017automatic}
A.~Paszke, S.~Gross, S.~Chintala, G.~Chanan, E.~Yang, Z.~DeVito, Z.~Lin,
  A.~Desmaison, L.~Antiga, and A.~Lerer, ``Automatic differentiation in
  pytorch,'' 2017.

\end{thebibliography}

%






\end{document}